# Physical Activation Functions (PAFs): An Approach for More Efficient Induction of Physics into Physics-Informed Neural Networks (PINNs)

*Jassem Abbasi* *, *Pål Østebø Andersen*

*Department of Energy Resources, University of Stavanger, 4036 Norway*

*\* Corresponding author: jassem.abbasi@uis.no*

**Abstract**

In recent years, the evolution of Physics-Informed Neural Networks (PINNs) has reduced the gap between Deep Learning (DL) based methods and analytical/numerical approaches in scientific computing. However, there are still complications in training PINNs and the optimal interleaving of physical models. These complications made PINNs suffer from inefficient extrapolation and slow training. In this work, we introduce the concept of Physical Activation Functions (PAFs). Instead of using standard activation functions (AFs) such as *ReLU*, *tanh*, and *sigmoid* for all the neurons, one can use generic AFs with their mathematical expression inherited from the physical description of the investigated phenomenon. The expression of PAFs could successfully be selected based on e.g., individual terms appearing in the analytical solution, the initial or boundary conditions of the PDE system. The PAFs can be applied in neurons of hidden layers in a PINN in combination with other AFs. We validated the advantages of PAFs in both forward and inverse problems for several PDEs, such as Burger's equation, the Advection-Convection equation, the heterogeneous diffusion equation, and the Laplace equation. The main advantage of PAFs was the more efficient constraining and interleaving of PINNs with the physical phenomena and their underlying mathematical models. The added PAFs significantly improved the predictions of PINNs for the testing data that were out-of-training distribution. Furthermore, applying PAFs reduced the size of the PINNs by up to 75% in different cases while maintaining the same accuracy. Also, the value of loss terms was reduced by one to two orders of magnitude in some cases, which is noteworthy for upgrading the training of the PINNs. It also improved the precision of the obtained results in inverse problems. It can be concluded that using the PAFs helps in generating PINNs with less complexity and much more validity for longer ranges of prediction. However, demonstration for more complex cases is required in the next studies.

# 1- Introduction

The capabilities of machine learning (ML) in different fields such as computer vision, and language processing have been proven in recent years. In consequence, the concept of scientific ML (SciML) has been introduced to take advantage of ML in scientific problems (Hey et al., 2020; Mjolsness & DeCoste, 2001). One of the main techniques of SciML is the Physics Informed Neural Networks (PINNs) that try to fuse the governing laws of the investigating systems (that may be in the forms of Ordinary Differential Equations (ODEs) or Partial Differential Equations (PDEs)) to Neural Networks (NNs). Actually, PINNs methodology tries to train NNs that can approximately reflect the physical behavior of the problem (usually in a spatial-temporal domain), including its initial and boundary conditions (Raissi et al., 2019). There is hope that progresses in the PINNs techniques will lead to more interpretable and reliable SciML models, especially in dealing with scientific problems where the accuracy and flexibility of the model are critical (Kochkov et al., 2021). The PINNs are trained by enforcing the NNs to follow the provided measured/calculated datasets and the mathematical representation of the involved physics. In recent years, many studies applied PINNs in different technical fields such as mechanics (Moseley et al., 2020), fluid dynamics (Cai et al., 2022; Jin et al., 2021), power systems (Misyris et al., 2020), and geosciences (Almajid & Abu-Al-Saud, 2022; Fuks & Tchelepi, 2020; He et al., 2020; Li et al., 2021).

In general, embedding physics into the NNs is performed by three pathways: observational data, physical loss functions, or mathematical induction (Karniadakis et al., 2021). Observational data is the traditional approach for training ML models in which the model is learned to accurately interpolate between the data that ought to reflect the underlying physical principles. This is the weakest form of interleaving the physics in the model. The second approach, physical loss functions, is added as a soft condition to the training algorithm to yield forecasts roughly satisfying a given set of physical constraints (for instance, their underlying differential equations). The mathematical induction, the most principled approach, fuses the physics by tailored interventions in the model architecture (Rodriguez-Torrado et al., 2021), commonly expressed in the form of certain mathematical constraints (Djeumou et al., 2021). However, the induction of physics into the DL models may apply only to some limited cases (Karniadakis et al., 2021). The mathematical proof of PINNs originates from the Universal Approximation Theorem (Hornik et al., 1989), which states that NN can closely approximate any continuous function with only one hidden layer and a finite number of neurons. However, although the Universal Approximation Theorem guarantees the existence of a neural approximator for the given mathematical problem, it does not offer any procedure for selecting the best network architecture or training the NN (Colbrook et al., 2022). Many researchers dealing with PINNs reported that minimizing the loss functions during a training operation is challenging, especially when the complexity of the corresponding differential equations increases (Leiteritz & Pflüger, 2021; Markidis, 2021; Shin et al., 2020). To deal with this difficulty, numerous approaches were suggested. Krishnapriyan et al. (2021) recommended a sequence-to-sequence practice where the data from the space-time domain is imported into the model partially during the training process. Also, they recommended the Curriculum Regularization approach where the complexity of the target loss function (e.g., the conditions that the PINN should meet) is increased during the learning.

Some of the studies on the improvement of PINNs focused on the modification of activation functions (AFs) of the neurons in the model. The choice of AFs in NNs may significantly impact training efficiency and model performance (Ramachandran et al., 2017). Different studies, such as Jagtap et al. (2020) and Jagtap et al. (2021), introduced the concept of a locally adaptive activation function where a scalable parameter is added to each layer/neuron that avoids the gradient descent optimization approaches attracted to local minimums. Various studies also tried to introduce special network architectures to combine different mathematical equations and NNs with each other (Lu et al., 2021; Peng et al., 2020; Shukla et al., 2021). To establish more physics-constrained NNs, some studies also suggested their co-utilization with symbolic regression approaches (Cranmer et al., 2020; Kubalík et al., 2021). However, there are still many challenges in developing authentic PINNs for scientific computations, particularly extrapolation to the out-of-training-distribution data (Kim et al., 2020).

In this work, a new concept called Physical Activation Function (PAF), which can be considered a mathematical induction approach to passing physics into ML models will be introduced. This approach helps make the PINNs more constrained to the expected physical/mathematical behavior. It will be shown that using this concept can substantially reduce the NNs size and improve the training process of PINNs. Also, it improves the extrapolation capabilities of PINNs in dealing with out-of-training-distribution data (i.e., domains that are out of training range). In the following, the theoretical background of the concept will be introduced, and then the advantages of the approach will be demonstrated by various examples.

## 2- Mathematical Definition

In scientific ML, we are interested in finding unknown variables from known variables. Here, we consider that the relationship between known vs. unknown variables is representable by differential equations (ODE or PDEs). Given $y$ as the unknown variable of interest as a function of known variables $x_1, \ldots, x_d$ and $t$, the mathematical equation of the physics is represented in the format of ODE or PDE, with the general form of:

$$\begin{aligned} \mathcal{F}(y, x; v) &= 0 \quad x \text{ in } \Omega, \\ \mathcal{B}(y, x; v) &= 0 \quad x \text{ in } \partial\Omega \\ \mathcal{I}(y, x; v) &= 0 \quad x \text{ in } \Omega, t = 0 \end{aligned} \quad (1)$$

Defined on the domain $\Omega \subset \mathbb{R}^d$ with the boundary domain, $\partial\Omega$ and $d$ are the spatial dimensions of the system. Here, $x := [x_1, \ldots, x_d; t]$ indicates the spatial-temporal coordinate vectors, $v$ is the function meta parameters related to the physics of the problem, and $\mathcal{F}$ is the differential operator. $\mathcal{I}$ and $\mathcal{B}$ denote the arbitrary initial and boundary (e.g., Dirichlet boundary) conditions of the problem, respectively. In the following, the mathematical representation of NNs and PINNs is introduced. These derivations aim to find a neural surrogate for the $\mathcal{F}$ is the differential operator. In the following, the Physical Activation Function (PAF) concept is introduced mathematically.

### 2-1- Overview of Neural Networks

A fully connected feedforward NNs, or a Multi-layer Perceptron (MLP), is a long sequence of units (neurons) connected with simple linear operations. In a generic NN called $\mathcal{N}$, there are $\mathcal{D}$ layers ($\mathcal{D} - 2$ hidden layers). In the $k_{th}$ layer, there are $n_k$ neurons. Each neuron receives an input, $x_{k-1}$, from the previous layer that is linearly transformed (Jagtap et al., 2020):

$$\ell(x_{k-1}) \triangleq w_k x_{k-1} + b_k; \quad k = 1, 2, \ldots, \mathcal{D} \quad (2)$$

The network weights ($w_k \in \mathbb{R}^{n_k \times n_{k-1}}$) and biases ($b_k \in \mathbb{R}^{n_k}$) are trainable parameters (Figure 1). The output of each neuron is passed through an activation function, $\sigma(\cdot)$, and then sent to the next layer as:

$$\ell(x_{k-1}) = \sigma(w_k x_{k-1} + b_k); \quad k = 1, 2, \ldots, \mathcal{D} \quad (3)$$

Then the final output of the NN is mathematically calculated from the inputs as:

$$\mathcal{N}(x;\theta) = \sigma(w_l \sigma(\cdots \sigma(w_2 \sigma((\omega_1 x) + b_1) + b_2) \cdots) + b_l) \quad (4)$$

where, $\theta = \{w_k, b_k\}$ represents the set of trainable parameters in the NN. The number of parameters in set $\theta$ is $N_\theta$. Using a different mathematical representation:

$$\mathcal{N}(x;\theta) = (\ell_D \circ \ell_{D-1} \ldots \circ \ell_1)(x) \quad (5)$$

where ∘ denotes the function decomposition, e.g., for sample functions $f(x)$ and $g(x)$ it is written as:

$$(f \circ g)(x) = f(g(x)) \quad (6)$$

Figure 1 shows the graphical representation of a fully connected feedforward NN. In the training of the NN with a set of $\{x, y\}$ dataset, the goal is finding a set of $\theta^*$ such that the defined loss function is minimized:

$$\theta^* = arg_\theta min \frac{1}{N} \sum_{j=1}^{N} \mathcal{L}(y_j, \mathcal{N}(x_j; \theta)) \quad (7)$$

where $N$ is the number of data points, and $\mathcal{L}$ is the loss function. Here, $\mathcal{L}$ is the mean squared error (MSE) between the true and predicted values:

$$\mathcal{L}(y_i, \mathcal{N}(x_i; \theta)) = \frac{1}{N} \sum_{i=1}^{N} (y_i - \mathcal{N}(x_i; \theta))^2 \quad (8)$$

The trainable parameters ($\theta$) are optimized to minimize the loss function in an iterative way. In each iteration (epoch), the θ set is updated as follows:

$$\theta^{m+1} = \theta^m - \eta_l \nabla_\theta \mathcal{L}(\theta) \quad (9)$$

where $m$ and $\eta_l$ denote the epoch number and the optimizer's learning rate, respectively. The training is carried out on training data, and to avoid underfitting or overfitting, the model is checked on separate validation data during the training process. The training process is stopped when the error of validation data starts increasing.

## 2-2- Overview of Physics Informed Neural Networks

A typical PINN tries to take benefits from observational data and physical loss functions. According to the Universal Approximation Theorem of NNs, for any sufficiently regular unknown variable $y$, there is a large enough NN, $\mathcal{N}$, with trainable parameters $\theta$, such that:

$$y(x) \approx y^*(x) = \mathcal{N}(x;\theta^*), \quad x \in \Omega \quad (10)$$

where $y^*(x)$ is the output of the NN. Given the known governing PDE in Eq. (1):

$$\begin{aligned} \mathcal{F}(y^*; v) &\approx 0 & x \text{ in } \Omega, \\ \mathcal{B}(y^*; v) &\approx 0 & x \text{ in } \partial\Omega \\ \mathcal{I}(y^*; v) &\approx 0 & x \text{ in } \Omega, t = 0 \end{aligned} \quad (11)$$

Eqs. (10 and (11 suggest that an arbitrary PDE can be solved by replacing the unknown $y(x)$ with a neural network $y^*(x) = \mathcal{N}(x;\theta)$ into the constituent PDE and finding the optimum trainable parameters $\theta^*$ such that $\mathcal{F}(\mathcal{N},v) \approx 0$, $\mathcal{B}(\mathcal{N};v) \approx 0$ and $\mathcal{J}(\mathcal{N};v) \approx 0$ (see Figure 1). The differential values of $\mathcal{N}$ versus the input parameters are calculated using Automatic Differentiation (AD) tools that are available in all modern DL libraries such as TensorFlow and PyTorch. Formally, we need to assess the validity of $\mathcal{N}$ in meeting the conditions in Eq. (11). So, a loss function called physical loss function ($\mathcal{L}_p$) is defined in the domain Ω:

$$\mathcal{L}_p(\theta,x) = \int_\Omega |\mathcal{F}(\mathcal{N},v)|^2 \, dx + \int_{\partial\Omega} |\mathcal{B}(\mathcal{N},v)|^2 \, dx + \int_\Omega |\mathcal{J}(\mathcal{N},v)|^2 \, dx \tag{12}$$

where we wish to find the $\theta^*$ which minimizes the $\mathcal{L}_p(\theta,x)$. This loss term enforces the NN into learning the physics of the problem. If $y(x) \approx \mathcal{N}(x;\theta^*)$ and $\mathcal{L}_p(\theta^*,x) \approx 0$, by definition, the NN is a solution to the differential equation, or a neural surrogate to the analytical model (Zubov et al., 2021). This NN is called Physics Informed Neural Network (PINN). In a PINN, the loss function can be defined as a combination of several losses:

$$\mathcal{L}(\theta^*,x) = (\mathcal{L}_1, \mathcal{L}_2, \dots, \mathcal{L}_n) \tag{13}$$

Each loss term represents different specifications of the system such as the deviation of the NN prediction from the dataset (Eq. (8)) or deviations from the mathematical representation of the physics and its associated boundary/initial conditions (Eq. (12)).

Figure 1 shows the way that the loss terms are used to train a symbolic PINN by the backpropagation algorithm (Eq. (9)). These loss terms are optimized together with weighted linearization (Raissi et al., 2017):

$$\mathcal{L}(\theta) = \sum_{j=1}^{n} \lambda_i \mathcal{L}_j(\theta) \tag{14}$$

where $\lambda_i$ denotes the weight factor of each loss function. There is no clear workflow for the selection of these weight factors; however, the distribution of datasets and fidelity of the model have determining roles in the selection of the optimum $\lambda_i$ values (Cuomo et al., 2022).

### 2-3- Physical Activation Functions

Suppose the solution of the non-linear differential operator (e.g., PDE) $\mathcal{P}(y,x) = 0$ can be expressed analytically or numerically with a sequence of functions $\mathcal{P}_1, \mathcal{P}_2, \dots, \mathcal{P}_k$ such that:

$$y(x) = \mathcal{P}_1 \circ \mathcal{P}_2 \circ \dots \circ \mathcal{P}_k(x) \tag{15}$$

Where *x* is the input variable(s), including both spatial and temporal variables, and *y* is the ground-truth unknown variable of interest. Remembering Eq. (10), for a series of collocation points $\{x,y\}$ in domain Ω, say that there is a NN or PINN, $\mathcal{N}$, where $y^*(x) = \mathcal{N}(x;\theta^*)$, such that:

$$y(x) \approx y^*(x) \quad for\ x \in \Omega \tag{16}$$

Here, $\mathcal{N}(x;\theta)$ would be the neural surrogate of $\mathcal{P}(y,x)$ in domain Ω. According to Eq. (5), $y^*(x) = \mathcal{N}(x;\theta^*)$ can be written as the below sequence of operations:

$$y^*(x) = \ell_1 \circ \ell_2 \circ \ldots \circ \ell_m(x) \tag{17}$$

Suppose $\mathcal{N}'(z, \theta)$ is a part of the (PI)NN $\mathcal{N}$ that, by definition, may include a single neuron or a set of neurons in the network. For simplicity, we assume that $\mathcal{N}'$ only has a single hidden layer.

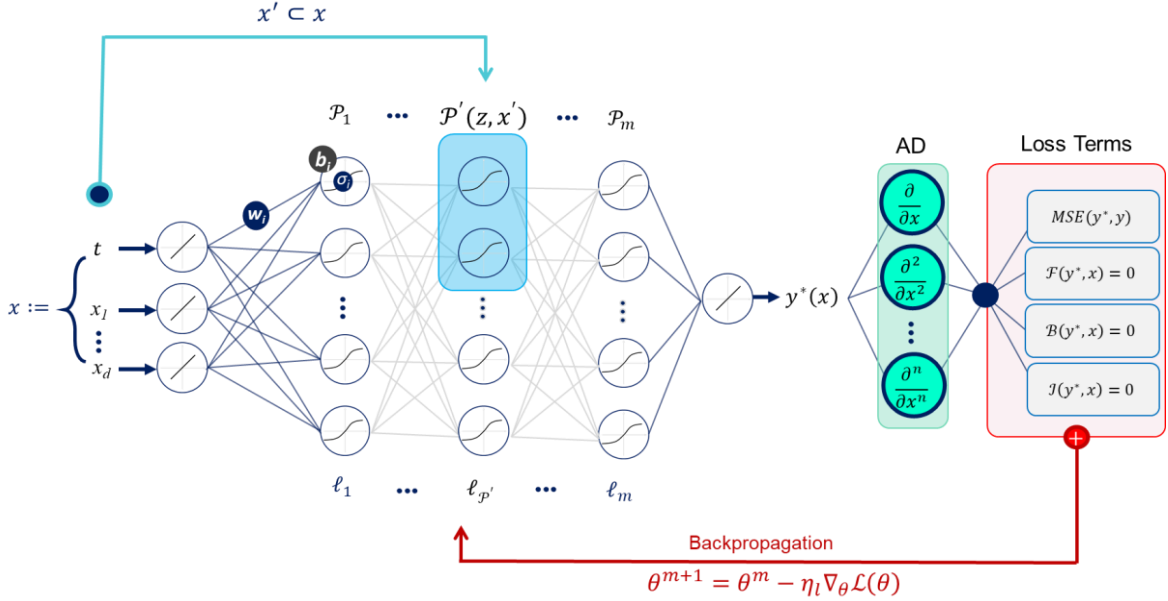

Figure 1: Simplified architecture of a PINN with inputs $x := [x_1, \ldots, x_d; t]$ and output $y^*(x)$. Here, the PINN is an MLP with $\ell_m$ layers. In this model, all the neurons in a layer may not have the same AFs. The differential of $y^*$ versus input variables are calculated using Automatic Differentiation (AD). The loss terms of PDE ($\mathcal{F}$) and initial ($\mathcal{I}$) and boundary ($\mathcal{B}$) conditions are calculated at each epoch. The trainable parameters are updated in a backpropagation algorithm using an optimizer. Here, $\mathcal{P}'(z, x')$ is an arbitrary PAF defined in the layer $\ell_{\mathcal{P}'}$. Also, $z$ is the input to the neuron from the previous layer, and $x'$ is a subset of the input variables.

Figure 1 illustrates how the $\mathcal{N}'$ may be placed in a (PI)NN. Here, $z$ is defined as the input values to the arbitrary hidden layer $k$ from the hidden layer $k$-1:

$$z = w^k \ell^{k-1} + b^k \tag{18}$$

According to the Universal Approximation Theorem of neural networks, suppose $\mathcal{N}'$ is responsible for regenerating a part of the operations required to solve $\mathcal{P}(y, x) = 0$, Eq. (15), called $\mathcal{P}'$ in domain $\Omega$. We can write:

$$\mathcal{N}'(z, \theta') \approx \mathcal{P}'(z, x') \tag{19}$$

where $\theta'$ is the optimum set of trainable parameters, and $x'$ is a subset of the input variables, i.e., $x' \subset x$. Suppose $\mathcal{P}'(z)$ is a generic continuous and preferably differentiable function inspired by the physical/mathematical model of the problem. We assume that for the operations carried out by $\mathcal{N}'(z, \theta')$ there is a replacement $\mathcal{P}'(z, x')$. The input of this function can be the output from the previous layer (Eq. (18)) or some of the input variables ($x'$). To make this new concept compatible with the current approaches in NNs, we implement this function into neurons of a NN as an activation function. We call this function a Physical Activation Function (PAF). In this work, we suggest substituting the operations carried out by the routine activation functions (e.g., *tanh*, and *sigmoid*) with the physical activation function $\mathcal{P}'(z, x')$. The functionality of a PAF may be inspired by the mathematical and/or physical behavior of the investigating phenomenon. If we replace the $\mathcal{P}'(z, x')$ into the hidden layer $\ell$, the operations carried out by $\mathcal{N}$ maybe redefiend as:

$$y^*(x) = \ell_1 \circ \ell_2 \circ \ldots \circ \ell_{\mathcal{P}'} \circ \ldots \circ \ell_m(x) \qquad (20)$$

where in this equation $\ell_{\mathcal{P}'}$ is the hidden layer that a PAF(s) has been implemented in that layer. The graphical representation of the PAF in a fully connected PINN is shown in Figure 1. As it is shown in this figure, PAFs can be implemented in parts of the neurons of a layer, not necessarily all of them. The PAFs can be considered a type of mathematical induction approach for more principal interleaving the physics into DL models. This work defines the fraction of the total neurons in the NNs using PAFs as the PAF fraction. Although we used a fully connected feedforward NN here, PAFs can be implemented in other network architectures, such as convolutional neural networks (CNNs) or recurrent neural networks (RNNs). The conditions for choosing the right PAFs are discussed in the next sections.

### 2-3-1- Training of PAF-PINNs

Considering that the backpropagation algorithm is the most determining tool in training the PINN, the impact of PAFs on the automatic differentiation process is also essential. The partial derivative of $y^*(x)$ versus $x$ in Eq. (20) gives:

$$\left.\frac{\partial y^*(x)}{\partial x}\right|_x = \left.\frac{\partial y^*}{\partial \ell_1}\right|_{\ell_1} \cdot \left.\frac{\partial \ell_1}{\partial \ell_2}\right|_{l_2} \cdot \ldots \cdot \frac{\partial \ell_{\mathcal{P}'}}{\partial z} \cdot \ldots \cdot \left.\frac{\partial \ell_m}{\partial x}\right|_x \qquad (21)$$

In this NN, the PAF $\mathcal{P}'$ is a part of the layer $\ell_{\mathcal{P}'}$. Since $\mathcal{P}'$ is inspired by the physical behavior of the investigating phenomenon, its derivative also can be considered physical. This is an advantage when we are dealing with automatic differentiation algorithms. Since $\partial \mathcal{P}'/\partial z$ is analytically representing the behavior of the system, less attempt is required for training based on the differentials of the PINN model versus input variables. It needs to be considered that the PAF needs to be continuous and differentiable in the investigating domain.

### 2-4- Implementation

All the mathematical models were implemented in Python 3.9.7 environment. The models were developed on top of the PyTorch 1.10 library as well as Pandas, Matplotlib, and NumPy libraries. Training of NNs was performed using Adam (Kingma & Ba, 2014) or LBFGS (Low-memory Broyden–Fletcher–Goldfarb–Shanno algorithm) optimizers (Sheppard et al., 2008) also to compare the performance of different optimizing algorithms. All the operations were performed on a personal laptop with an Intel Core i9 processor, and 32 GB RAM.

### 3- Results

In the following, the concept of Physical Activation Function is examined in different case studies. In these cases, generally, two loss terms were used. The basic loss term, named 'data-based loss' is the MSE of the NN (or PINN) predictions vs. the expected true values, i.e., Eq. (8). The second set of loss terms is called the 'physical loss', where the discrepancy between the governing equation (plus any acting constraints, e.g., initial conditions or boundary conditions), and the PINN approximation is evaluated and minimized during the training. The mathematical expression of this loss term is shown in Eq. (12). So, for PINN $\mathcal{N}$ and the collocation points $\{x, y\}$, with the mathematical expression $\mathcal{F}(x, y)$, boundary condition $\mathcal{B}(x, y)$, and initial condition $\mathcal{J}(x, y)$, the 'total loss' term is expressed as:

$$\mathcal{L}(\theta) = \frac{1}{N}\sum_{i=1}^{N}(\mathcal{N}-y)^2 + \frac{1}{N}\sum_{i=1}^{N}|\mathcal{F}(x,\mathcal{N})|^2 + \frac{1}{N}\sum_{i=1}^{N}|\mathcal{B}(x,\mathcal{N})|^2 + \frac{1}{N}\sum_{i=1}^{N}|\mathcal{J}(x,\mathcal{N})|^2 \tag{22}$$

where in the right-hand side of this equation, the first term is the data-based loss ($\mathcal{L}_d$), and the summation of the following terms is the physical loss ($\mathcal{L}_p$) that are expressions for the squared errors of the PDE, boundary conditions, and initial conditions, respectively. In the following section, the impact of PAFs on forward problems is studied.

## 3-1- Forward Problems

### 3-1-1 Harmonic Oscillator

Here we show the potential of using PAFs in a simple case with an analytical solution, i.e., a mechanical harmonic oscillator. Also, an analysis of the implementation of PAFs was carried out. The dynamics of movements in mechanical oscillators such as springs or pendulums is a classic problem of interest in mechanics. The position ($x$) of the oscillating object with respect to time ($t$) is written as:

$$f(x,t) = m\frac{\partial^2 x}{\partial t^2} + \mu\frac{\partial x}{\partial t} + kx = 0 \tag{23}$$

where $m$ is the mass, $k$ the spring force constant, and $\mu$ a constant that parameterizes the strength of the damping. This PDE has an exact analytical solution for a damped oscillation ($\delta < \omega_0$):

$$x(t) = e^{-\delta t}(2A\cos(\phi + \omega t)); \quad \text{with } \omega = \sqrt{\omega_0^2 - \delta^2} \tag{24}$$

when $x(t = 0) = 1$. Here, $\phi = \tan^{-1}(-\delta/\omega)$, $\delta = \mu/2m$, $\omega_0 = \sqrt{k/m}$, and $A = 1/2\cos(\phi)$.

We train an NN that can regenerate the solution of the above PDE (its exact analytical form). To do so, we prepared a set of collocation points $\{t, x\}$, and a physical loss function related to the PDE. Two loss functions of $\mathcal{L}_d$ and $\mathcal{L}_p$ were combined linearly, and the total loss function was imported to an Adam optimizer. In this example, the exact analytical solution of the PDE is available. As Eq. (24) shows, the mathematical behavior of a generic harmonic oscillator is analogous to the multiplication of $\cos x$ and $e^x$ functions. Based on the analytical solution, two PAFs were defined.

$$\sigma_{cos}(z) = \cos(z) \tag{25}$$

$$\sigma_{exp}(z) = \exp(z) \tag{26}$$

In this context, $z$ is the neuron's input from the posterior layer.

Before investigating the impact of PAFs, we tested the ability of PINNs in the accurate prediction of harmonic oscillation versus time, especially for the out-of-training datasets. Figure 2 shows the results.

The dashed lines are from the analytical model. The purple points are training data used for the minimization of $\mathcal{L}_d$ (data-based loss). The region where each loss term was active is shown in each plot in Figure 2. The output of the NN for the training range is shown in blue, while the testing data (out of range of the training data) is shown in red. The structure of the used NN and AFs used for their neurons are shown in each figure. In all the NNs, different types of AFs (i.e., *linear* and *tanh* AFs) were used.

Figure 2a and b show the NN output for two different training ranges; neither case used the physical loss function during training. In both cases, the NN could follow the correct trend only in the training range. In Figure 2c, a PINN model is utilized. Data were available in the range of 0-0.75 *s*, and $\mathcal{L}_p$ was added in the range 0-1.5 s. Adding $\mathcal{L}_p$ could improve the predictions in the range of 0.75-1.50 *s*, where the data was not available (in comparison to Figure 2a). This indicates that PINN can compensate for the lack of data in the training range. However, the main question is whether the trained PINN can also work well for the out-of-training-distribution ranges of data. For all the cases in Figure 2, the models failed even in predicting the trend of the testing data, demonstrating that adopting PINNs in traditional PDE solvers still requires more attention.

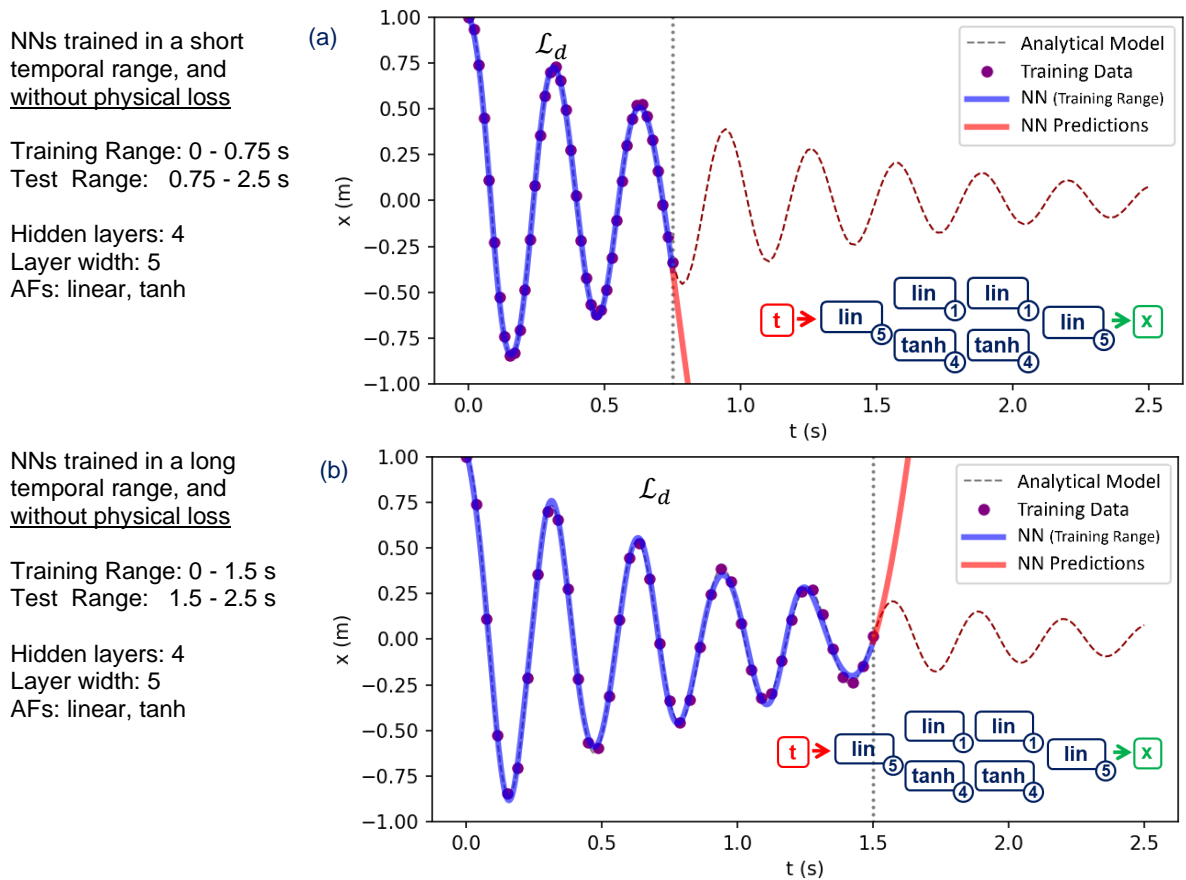

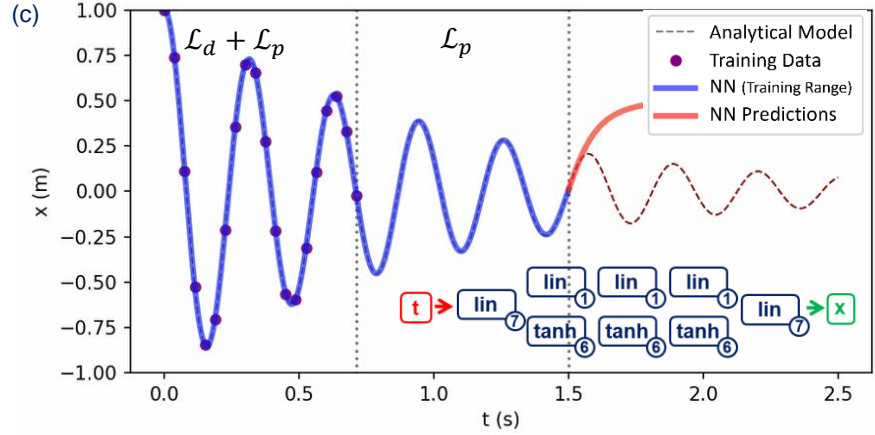

PINNs trained with a physical loss.

Training Range:
Data-Oriented: 0 – 0.75 s
Physics Oriented: 0 - 1.5 s

Test Range: 1.5 - 2.5 s

Hidden layers: 5
Layer width: 7
AFs: linear, tanh

*Figure 2: The capabilities of NNs and PINNs in the prediction of the harmonic oscillation out of the training range. The architecture of the NNs is shown in each case (for simplicity, the connections between neurons are hidden). Purple points are the training data (labeled as 'Data'), and the dash-line is the analytical solution of the PDE. The range that the physical loss is used is labeled as 'Physics. The output of NN in the training range is shown in blue, and its predictions for the testing data are shown in red.*

Then, we tried to assess the performance of the PAFs defined in Eqs. (30 and (31. The data ranges are the same in the following cases, summarized in Table 1 and Figure 4. As it is shown in Table 1, using PINNs instead of traditional NNs could reduce the total loss value ($\mathcal{L}_d + \mathcal{L}_p$) from 6.9e-4 to 4.8e-4. However, for the testing data (out-of-training-distribution prediction), the $R^2$ was negative for both cases (far from the expected values). For PINNs with PAFs, the loss value was reduced by two orders of magnitude, and the $R^2$ of testing data was improved significantly towards 0.97. The comparison of Figure 2 and Figure 3 shows these differences clearly.

Furthermore, by a trial-and-error approach, we searched for the smallest possible size of NNs for each case. Our observations showed that when a PAF was added to the network, the model could be trained with fewer parameters (in this example, $\dot{\theta}$ reduced from 246 to 67). These results indicate that smaller NNs with fewer trainable parameters are possible to be used when a PAF is included in the NN.

*Table 1: A summary of the investigated scenarios with different loss-function and activation functions for the mechanical harmonic oscillator PDE.*

| PINNs | PAF | Activation Functions | Training range (s) | | Total Loss ($m^2$) | Trainable Parameters ($\boldsymbol{\theta}$) | $R^2$ | |
| --- | --- | --- | --- | --- | --- | --- | --- | --- |
| | | | **Data-based** | **Physics-based** | | | **Train** | **Test** |
| **No** | No | Linear, tanh | 0.0 - 1.5 | - | 6.9e-4 | 106 | 0.99 | -63.74 |
| **No** | No | Linear, tanh | 0.0 - 0.75 | - | 2.5e-4 | 106 | 0.99 | -1230.2 |
| **Yes** | No | tanh | 0.0 – 0.75 | 0.0 - 1.5 | 4.8e-4 | 246 | 1.00 | -28.12 |
| **Yes** | Yes | tanh, $\sigma_{cos}$, $\sigma_{exp}$ | 0.0 – 0.75 | 0.0 - 1.5 | 5.2e-6 | 67 | 1.00 | 0.98 |
| **Yes** | Yes | tanh, $\sigma_{cos}$ | 0.0 – 0.75 | 0.0 - 1.5 | 6.9e-6 | 67 | 1.00 | 0.97 |

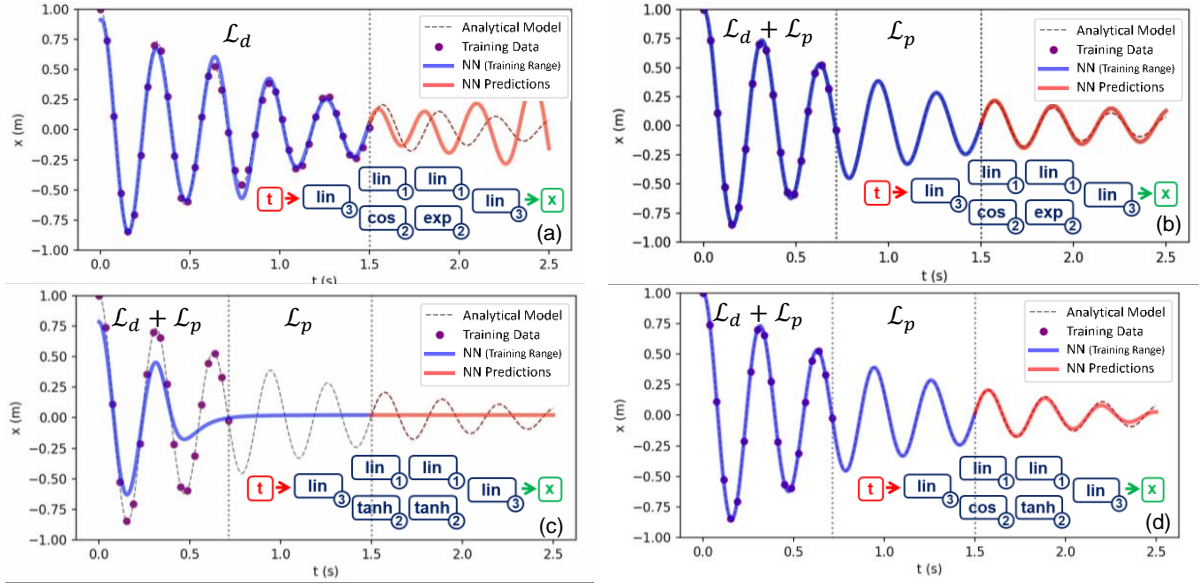

*Figure 3: The impact of PAFs on the predictions of NNs in the different scenarios. The used NNs are shown in each case (for simplicity, the connections between neurons are hidden). Purple points are the training data (labeled as 'Data'), and the dash-line is the analytical solution of the PDE. The range that the physical loss was used is labeled as 'Physics'. The blue line shows the training range, and the red line shows the NN predictions on the test data. a) NNs with $\sigma_{cos}$, and $\sigma_{exp}$ PAFs, and without physical loss; b) PINN with $\sigma_{cos}$, and $\sigma_{exp}$ PAFs, and with physical loss. c) PINN without PAFs and shallow hidden layers; d) PINN with $\sigma_{cos}$ PAF, and with physical loss.*

### 3-1-2 Advection-Dispersion Equation

In this case, we show the applicability of PAFs when only a limited part of the governing physics is known. Also, a comparison of the mixed usage of general AFs and PAFs is carried out. Hydrodynamic dispersion in a semi-infinite porous media is a common phenomenon in many subsurface processes, such as solute transport in aquifers and chemical flooding in oil recovery. The below PDE is yielded by combining advection and dispersion transport processes (one-dimensional and homogeneous system).

$$\frac{\partial C}{\partial t} = D \frac{\partial^2 C}{\partial x^2} - v \frac{\partial C}{\partial x} \qquad (27)$$

where *C* is the concentration of particles (*mol/kg*), *v* is fluid flow velocity (*m/s*), and *D* is the hydrodynamic dispersion coefficient (*m²/s*). Zheng & Bennett (2002) provided an analytical solution for this PDE:

$$C(x,t) = \frac{C_0}{2} \left[ erfc\left(\frac{x-vt}{2\sqrt{Dt}}\right) + \exp\left(\frac{vx}{D}\right) erfc\left(\frac{x+vt}{2\sqrt{Dt}}\right) \right] \qquad (28)$$

by the condition that:

$$\begin{array}{ll} C(x,0) = 0 & x \geq 0 \\ C(0,t) = C_0 & t \geq 0 \\ C(\infty,t) = 0 & t \geq 0 \end{array} \qquad (29)$$

In this case, we intend to train a PINN that can predict the advection-dispersion phenomena as a surrogate model for Eq. (27). Based on the analytical solution, we defined two different physical activation functions as below:

$$\sigma_{erfc}(x, z) = erfc\left(\frac{x - z}{2\sqrt{z}}\right) \quad (30)$$

$$\sigma_{exp.erfc}(x, z) = \exp(x)\, erfc\left(\frac{x + z}{2\sqrt{z}}\right) \quad (31)$$

where $z$ is the input to the neuron, and $x$ is the position. The data required for training the PINN is gathered for $C_0=1$, $v=0.003$, and $D=0.003$. The data was collected in the $t$ range of 0-300 $s$, and the $x$ range of 0-2 mm. The complete details about the ranges of data are provided in Table 2. In this example, as in previous examples, two data-based and physical losses were defined.

*Table 2: A summary of the ranges of the dataset used for training and testing the PINN model of the advection-convection phenomenon.*

| Data | x-range | t-range |
|---|---|---|
| **Data-based loss** | 0.0 – 0.5 | 0.0 – 50 |
| **Physics-based loss** | 0.0 – 1.0 | 0.0 – 100 |
| **Testing data** | 0.0 – 2.0 | 0.0 – 300 |

The results of the analysis are shown in Figure 4. For the base case (w/o PAFs), a PINN with a network size of $\dot{\theta} = 781$ was defined. All of the neurons used *tanh* AF. As Figure 4a depicts, the PINN model could perform acceptably in the training range but failed in the testing range. This is while a wide network with 781 trainable parameters was used. As a replacement, we generated a PINN model with a single hidden layer and only two neurons in that layer. We added the defined PAFs (Eqs. (30 and (31) to each of the neurons. The results after training the model are shown in Figure 4b. As it is clear, with only two neurons, the model could follow the real data exactly, and the exciting point is that the optimizer could find optimum weights. The obtained results were almost expected because we passed the true solution of the PDE to the PINN and induced it to follow the mathematical behavior of the phenomenon. However, this result is an indication of the theoretical advantage of using PAFs. To extend results to more practical cases, we only tested one of Eqs (30(31 in the PINNs (only a part of the behavior of the problem is known). As Figure 4c&d show, these models could act better in comparison to Figure 4a, while $N_{\dot{\theta}}$ reduced from 781 to 49.

However, since we used a mixed combination of AFs in the layers of NNs, one may be concerned about how this mixed usage of AFs may affect the PINNs performance. In other words, it is necessary to distinguish the impacts of using a mixed combination of AFs and PAFs. To do so, we analyzed the different combinations of AFs in a PINN model with four layers, each with ten neurons. The 2[nd] layer of the PINN is altered in various scenarios as 1) all of the neurons used tanh AF, 2) used a combination of general AFs such as *tanh*, *linear*, *ReLU*, *softmax*, and *sigmoid* AFs, 3) a combination of general AFs and PAFs, and 4) only used tanh and relevant PAFs. The results of the different scenarios are shown in Table 3. It is shown that using a mixed combination of general AFs (that are not specific to the physics of the problem) may reduce the efficiency of the PINN model. However, as PAFs were added to the investigating layer, the total loss terms were reduced, and $R^2$ values for the test data were improved.

Table 3: Comparing the usage of different AFs in the PINN. In all cases, four hidden layers with ten neurons were used (all with tanh AFs). In different cases, the AFs of the $2^{nd}$ layer was changed in different combinations. All scenarios include 481 trainable parameters and an LBFGS optimizer was used for the training. The number in brackets shows the number of neurons for the AF.

| Activation Functions [number of neurons] | Total Loss | $R^2$ | |
|---|---|---|---|
| | | Train | Test |
| tanh[10] | 6.7e-5 | 0.995 | 0.672 |
| tanh[2], linear[2], sigmoid[2], softmax[2], ReLU[2] | 9.7e-5 | 0.996 | 0.413 |
| tanh[5], $\sigma_{exp}$[1], $\sigma_{erfc}$[1], sigmoid[1], softmax[1], ReLU[1] | 5.3e-5 | 0.997 | 0.871 |
| tanh[6], $\sigma_{exp.erfc}$[2], $\sigma_{erfc}$[2], | 5.2e-5 | 0.995 | 0.942 |

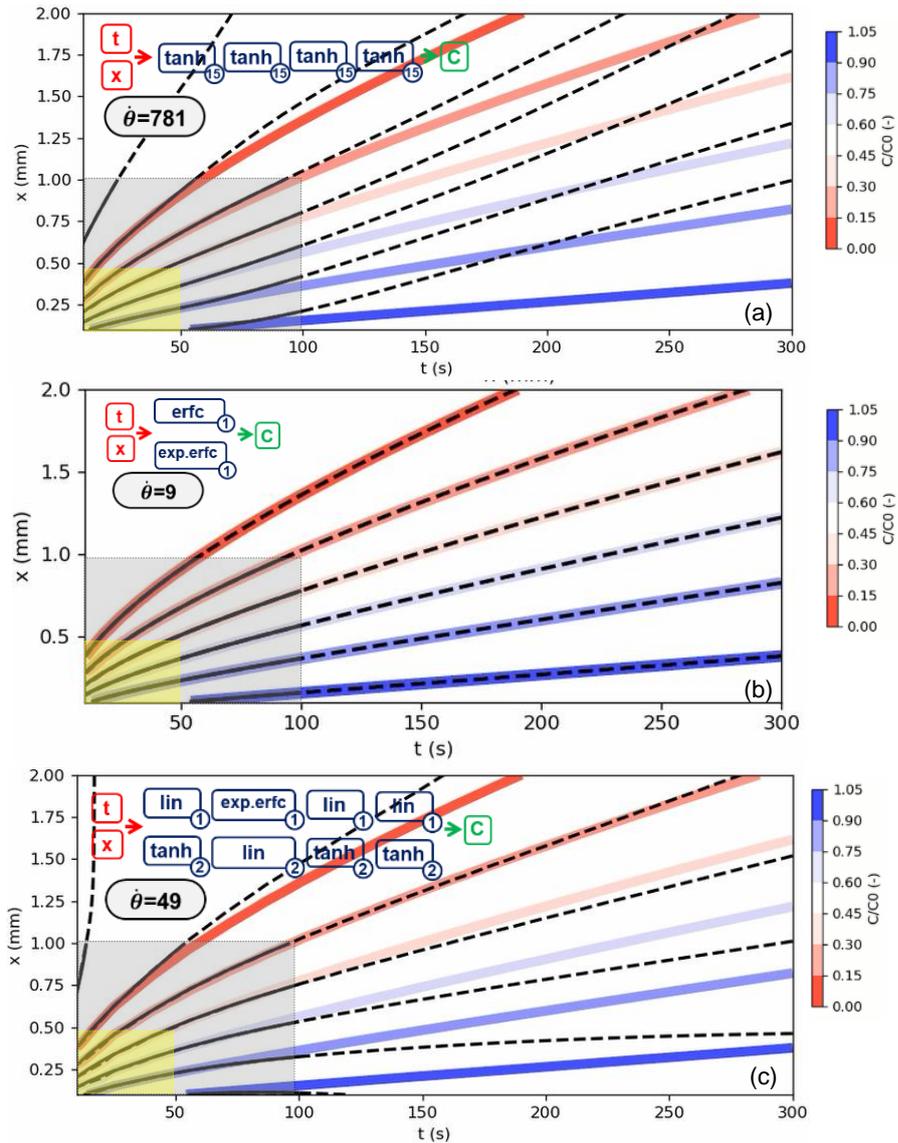

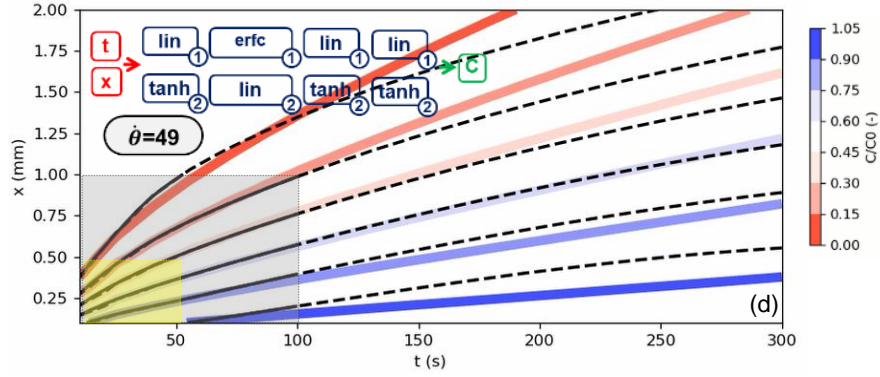

*Figure 4: Comparison of implementing different AFs in the PINN for prediction of concentration ($C/C_0$) in the advection-dispersion equation. In all the figures, the yellow/grey area in all the figures shows the data/physics-based training range, respectively, while the remaining area is related to the testing data. The NN structure and the number of trainable parameters ($\dot{\theta}$) are shown in each figure. Black solid/dashed lines represent the NN output for training/testing regions, respectively, while the colored contour map shows the true values of dimensionless concentration.*

### 3-1-3 Diffusion Equation

In many practical cases, the analytical solution may not be available easily. In this case, we try to see if it is possible to use any known other functionalities between the variables as a PAF. To do so, we considered a 1D transient diffusion phenomenon with PDE:

$$\frac{\partial(C)}{\partial t} = \frac{\partial}{\partial x}\left(D\frac{\partial C}{\partial x}\right) \tag{32}$$

Where $C$ is the concentration of particles (*mol/kg*), and $D$ is the diffusion coefficient ($m^2/s$). We numerically solved this diffusion problem using the *Py-PDE* library (Zwicker, 2020). Using this module, we were able to choose arbitrary functions for both the initial distribution of concentration and diffusion coefficient. So, we assumed that the diffusion coefficient changes versus spatial coordinate as:

$$D(x) = 0.02(0.3x + \log(x^2 + 1.5)) \tag{33}$$

With initial condition:

$$C(x, t = 0) = 0.2\left(\frac{1}{x^2 + 0.2}\right)^2 \tag{34}$$

And boundary conditions:

$$\begin{aligned} C(x = -\infty, t) = 0 \quad t \geq 0 \\ C(x = +\infty, t) = 0 \quad t \geq 0 \end{aligned} \tag{35}$$

These conditions create a 1D diffusion from the point $x = 0$ to the surrounding area. The diffusion coefficient also changes versus *x*. This problem does not have any analytical solutions.

The LBFGS optimizer is used in this example. To train a PINN for this problem, same as in the previous cases, two different losses were defined (data-based loss and physics-based loss). Data from the numerical simulation was exported in $x$ range of $[-1,1]$ *mm* and *t* range $=[1,30]$ *s,* with a spatial resolution of 0.1 *mm* and temporal resolution of 1 *s*. The PINN model was trained based on the data in the time range of [0,5] *s* and based on physics in the time range of 0-10 *s*. In both loss terms, the *x* range was -1 to 1 *mm*. To restrict the PINN training for the times beyond the training range, the function of *D* versus *x* and the initial condition of the above problem were converted to PAFs as below:

$$\sigma_{logx2}(z) = (0.3z + \log(z^2 + 1.5)) \tag{36}$$

$$\sigma_{invx2}(z) = 0.2(\frac{1}{z^2 + 0.2})^2 \tag{37}$$

Then, we investigated the impact of using the PAFs in PINNs with different network widths. The difference between the output of PINN with and w/o PAFs is shown in Figure 5 and Table 4. The architecture of the PINN is shown in Figure 5. As the table depicts, using the PAFs could increase the $R^2$ of the test data for both network widths. This is while for the widest network, the loss function decreased. This example reveals the potential of adding initial or boundary conditions in an explicit form to the NNs model using the concept of PAFs. This potential helps enforce the PINN to follow the expected physical behavior explicitly, even in the untrained regions.

Table 4: *A summary of the investigated scenarios with different activation functions for the dynamic diffusion PDE with arbitrary initial conditions and diffusion coefficient function.*

| Activation Functions | PAF | Trainable Parameters | Loss | | | $R^2$ | |
|---|---|---|---|---|---|---|---|
| | | | Total | $\mathcal{L}_d$ | $\mathcal{L}_p$ | Train | Test |
| *tanh* | No | 111 | 6.9e-4 | 6.5e-6 | 1.9e-4 | 1.0 | 0.972 |
| *tanh*+$\sigma_{logx2}$ + $\sigma_{invx2}$ | Yes | 111 | 7.7e-4 | 7.3e-6 | 2.3e-4 | 1.0 | 0.995 |
| *tanh* | No | 371 | 1.3e-4 | 6.3e-7 | 8.5e-5 | 1.0 | 0.927 |
| *tanh*+$\sigma_{logx2}$ + $\sigma_{invx2}$ | Yes | 371 | 8.7e-5 | 6.5e-7 | 3.9-e5 | 1.0 | 0.994 |

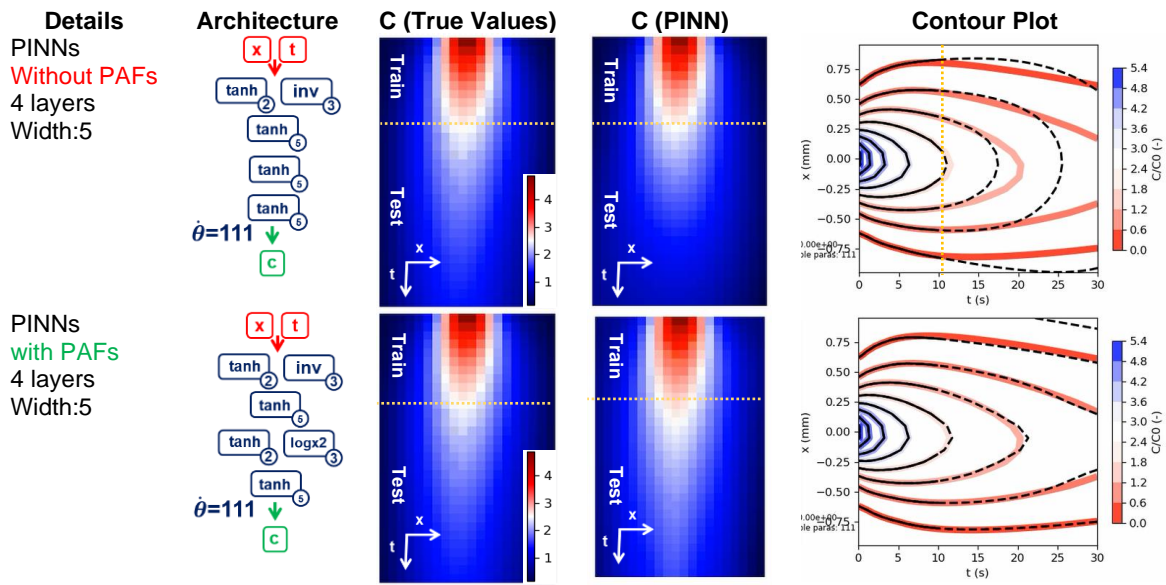

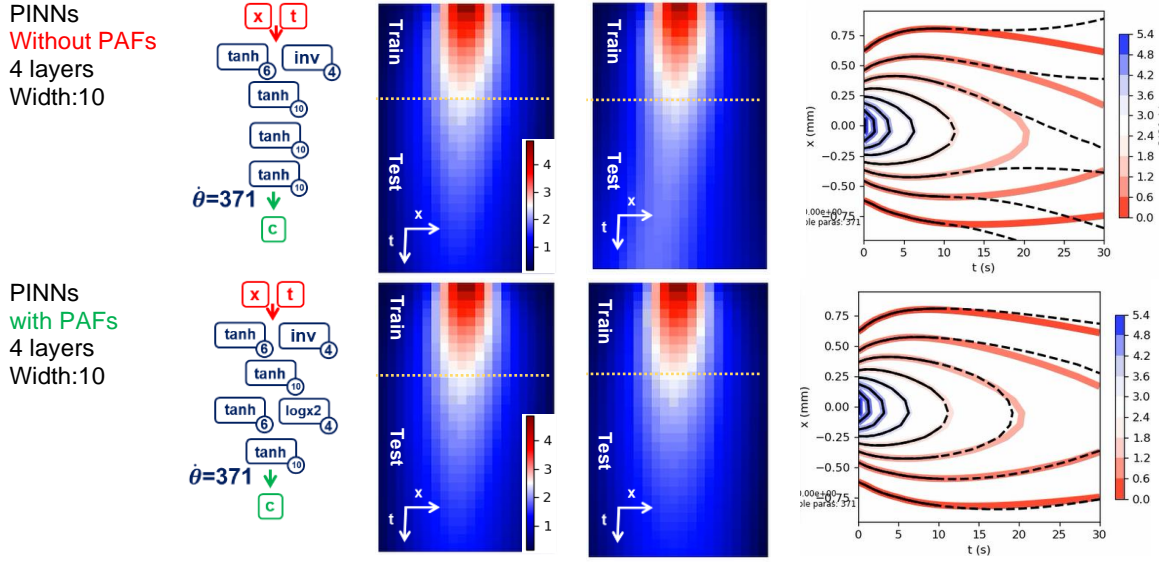

*Figure 5: 1-dimensional diffusion when D is a function of position; Comparison of the performance of PINNs before and after implementing PAFs. The first column in each row shows the details of each case. In the contour maps, the colored contour lines show the true values, the black lines show the PINNs outputs, the solid line is used on the training data range, and the dashed line is for the test data range.*

In the following example, to extend the complexity of the problem, we held the initial concentration as Eq. (34) but altered the diffusion coefficient function from Eq. (33) to:

$$D(x) = 0.02 * \exp\left(-(x + \sin(0.5t))^2\right) \quad (38)$$

This equation makes the diffusion coefficient periodically dependent on time. The above equation lets us define the below PAFs:

$$\sigma_{\text{expx}}(x, z) = 0.02 * \exp\left(-(x + \sin(z))^2\right) \quad (39)$$

$$\sigma_{expt}(t, z) = 0.02 * \exp\left(-(z + \sin(t))^2\right) \quad (40)$$

In these equations, we derived two different PAFs from Eq. (38), where in each of them, *t* and *x* were replaced by *z*. We then imported these PAFs into a PINN structure. The PINN was trained in the time range of 0-15 s and tested in the time range of 15-30 s. The model was trained with the LBFGS optimizer. The results are comparable in Table 4 and Figure 6. It can be seen that the accuracy of the PINN predictions for the testing data was improved significantly. As depicted in Figure 6, the PINN with PAFs could approximately follow the trend of diffusion versus time, while for the PINN case without PAFs, the predictions significantly deviated from the ground truth data, except in the training range. From Table 4, it can be concluded that using PAFs could help the optimizer reduce the physics-based loss function.

*Table 5: A summary of the investigated scenarios with different loss-function and activation functions for Burger's equation with the periodic initial condition.*

| Activation Functions | Loss | Trainable Parameters | Total Loss | Loss Data | Physics Loss | $R^2$ train | $R^2$ total |
|---|---|---|---|---|---|---|---|
| *tanh* | PINN | 481 | 2.1e-3 | 5.3e-4 | 1.5e-3 | 0.999 | 0.878 |
| *tanh*+$\sigma_{logx2} + \sigma_{invx2}$ | PINN | 481 | 2.1e-3 | 6.6e-4 | 1.3e-3 | 0.999 | 0.994 |

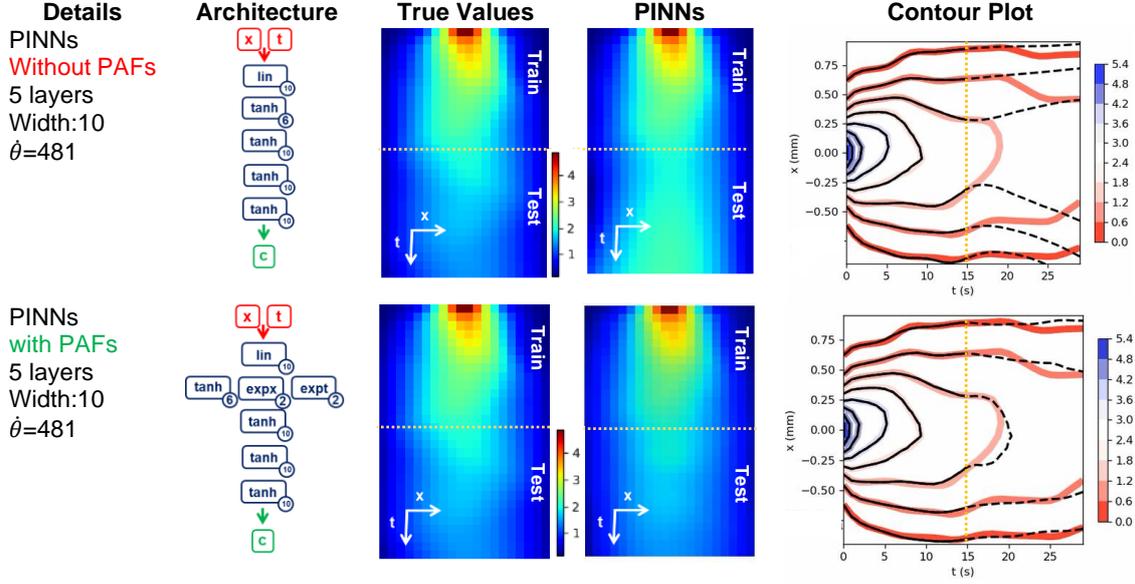

*Figure 6: 1D diffusion when the diffusion coefficient (D) is a function of both time and position; Comparison of the performance of PINN before and after implementing PAFs. The first column in each row shows the details of each case. In the contour maps (last column), the colored contours show the true values, the black lines show the PINN outlets, the solid line is for the training data, and the dashed line is for the test data.*

### 3-2 Inverse Calculations

This section examines the application of PAFs in improving two inverse PINN problems, including the diffusion equation and the 2D Laplace equation. We used the LBFGS optimization algorithm in both cases. Like in the previous examples, the network was an MLP; and to make the instances comparable (avoid initialization biases), we initialized the trainable parameters once and applied them to the different scenarios.

### 3-2-1 Diffusion Equation

In this case, the diffusion equation is studied with the initial conditions like Eq. (34) and the domain diffusion coefficient as a function of $x$:

$$D(x) = 0.01(0.30x^2 + 0.30) \tag{41}$$

As an inverse problem to estimate this function $D(x)$, we assumed that the coefficient in this equation is unknown as:

$$D(x) = 0.01(\lambda_1 x^2 + \lambda_2) \tag{42}$$

where true $\lambda_1$ and $\lambda_2$ are both equal to 0.30. So, in this case, we defined a PAF as:

$$\sigma_D(z, \omega_1, \omega_2) = 0.01(\omega_1 z^2 + \omega_2) \tag{43}$$

where $\omega_1$ and $\omega_2$ are also trainable parameters. In this experiment, we defined an MLP network with trainable parameters $\theta = \{w, b, \lambda, \omega\}$. We extracted a set of data from numerical simulations in the range of x= $(-1, +1)$ m and $t = [0, 40]$ s. In total, 82 data points were randomly chosen from the numerical simulation data to be used in the inverse calculations. The used MLP was a 4-layer network with a depth of 9 neurons. Three different networks are investigated below:

1. **PAF_0**: An MLP with *tanh* activation function in all neurons, without PAF
2. **PAF_2**: An MLP with *tanh* activation function, except two neurons with PAF (Eq. (43)) in the second layer
3. **PAF_4**: An MLP with *tanh* activation function, except four neurons with PAF (Eq. (43)) in the second layer

Each of the networks had 305 trainable parameters. The calculations have been performed with the LR of 0.003 and 0.001, each in 8000 epochs. The obtained results are summarized in

Table 6 and *Figure 7*. As seen in *Figure 7*, the models with PAF had lower loss values. Also, the errors in the prediction of $\lambda_1$ and $\lambda_2$ reduced significantly by including PAF in the MLP model.

Table 6: The results of inverse calculations using PAFs used for obtaining the values of $\lambda_1$ and $\lambda_2$.

| MLP | Loss | $\lambda_1, \lambda_2$ | ARE $\lambda_1, \lambda_2$ |
|---|---|---|---|
| PAF_0 | 1.59e-04 | 0.322, 0.288 | 0.073, 0.040 |
| PAF_2 | 7.41e-05 | 0.306, 0.289 | 0.020, 0.036 |
| PAF_4 | 1.19-04 | 0.305, 0.289 | 0.016, 0.036 |

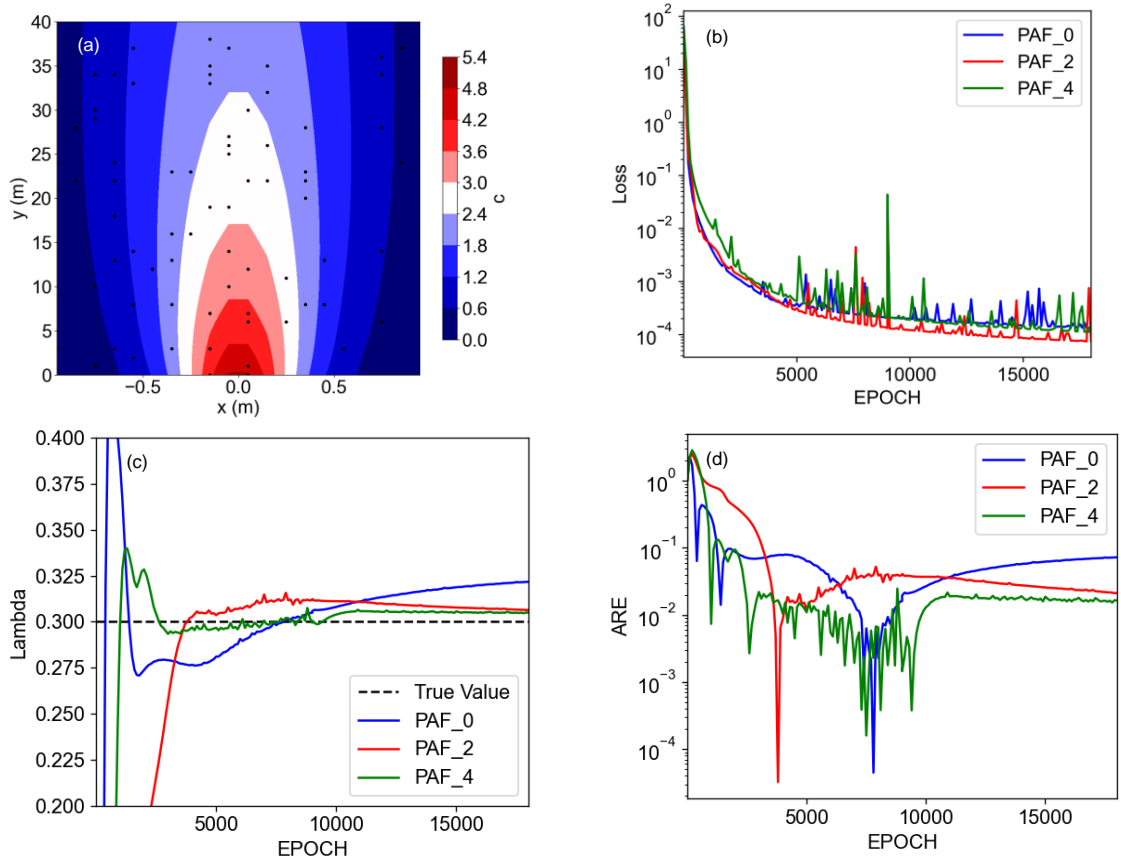

*Figure 7: The results of using PINNs for inverse calculation of diffusion coefficient parameters with and without using PAFs. a) The contour map of the diffusion process and the randomly chosen data (black dots) in the domain. a) the optimizing process done in 16000 epochs, with LR of 0.003 and 0.001, b) the value of $\lambda_1$ during optimization, c) the ARE values for the cases with and without PAF.*

### 3-2-2 Laplace Equation

In this second inverse problem, the Laplace equation is studied, where the goal is to calculate unknown parameters of the periodic boundaries in the Cartesian scale. The Laplace Equation is defined as:

$$\nabla^2 u = \frac{\partial^2 u}{\partial x^2} + \frac{\partial^2 u}{\partial y^2} = 0 \qquad (44)$$

This equation is solvable by constraining the boundaries. Here, the boundaries of the system were constrained to:

$$u(x, y, \lambda_1) = 3y \sin(y + \lambda_1); \quad \text{for} \quad x = 0, 2\pi \qquad (45)$$

$$u(x, y, \lambda_2) = 3x \sin(x + \lambda_2); \quad \text{for} \quad y = 0, 2\pi \qquad (46)$$

In this problem, $\lambda_1$ and $\lambda_2$ are unknown variables whose real values need to be estimated. We generated a dataset using numerical simulation (with true values $\lambda_1 = \lambda_2 = 3.14$). From this dataset, as the observation data, we randomly separated a set of collocation points in the finite-dimensional space and on the boundaries (*Figure 8*a). By considering Eqs. (45 and (46, here we defined the PAF as below:

$$\sigma_D(z, \omega_1) = 3z \sin(z + \omega_1); \qquad (47)$$

where z is the output of the neuron and the $\omega_1$ is the trainable parameter that is changed during the training of the PINN. To find the values of $\lambda_1$ and $\lambda_2$ by inverse calculation, different scenarios are evaluated using Eq. (47 in the second layer of the network. The training processes are shown in Figure 8b, where NNs with 1 or 2 PAFs achieved lower loss terms. The calculated $\lambda_1$ values and its relative error are shown in Figure 8c and d, and in Table 7, where the NNs with 1-3 neurons resulted in more accuracy. This example also shows how PAFs could improve the inverse calculations by constraining the NN and the automatic differentiation approach. However, the dominance of PAFs in a single layer of NNs can reduce flexibility and weaken the calculations.

*Table 7: The results of inverse calculations used for obtaining the values of $\lambda_1$ and $\lambda_2$ in the PINNs with different numbers of PAFs.*

| MLP | Loss | $\lambda_1, \lambda_2$ | ARE $\lambda_1, \lambda_2$ |
|---|---|---|---|
| PAF_0 | 13.8 | 2.83, 3.01 | 0.099, 0.041 |
| PAF_1 | 4.61 | 3.12, 3.14 | 0.007, 0.001 |
| PAF_2 | 4.71 | 3.08, 3.14 | 0.018, 0.001 |
| PAF_3 | 2.57 | 3.12, 3.22 | 0.005, 0.027 |
| PAF_4 | 7.04 | 2.90, 1.43 | 0.075, 0.544 |
| PAF_5 | 2.39 | 2.65, 2.94 | 0.153, 0.062 |

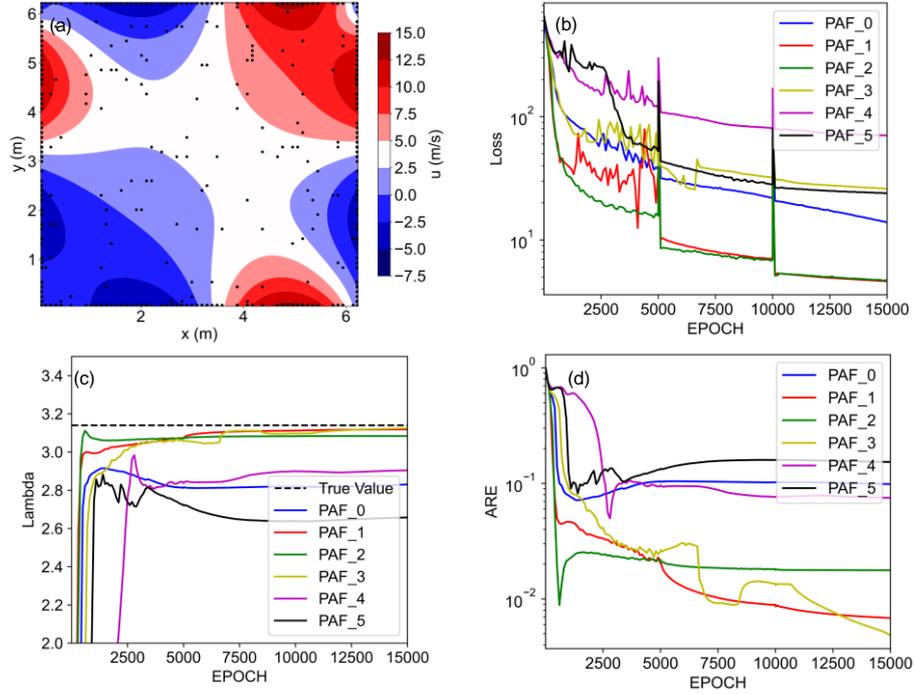

*Figure 8: Utilizing PINNs for inverse calculation of boundary condition parameters in the Laplace equation. a) The contour map of the system in 2D and the randomly chosen points in the domain and on the boundaries, b) the training process performed in 15000 epochs, with LR of 0.003 to 0.001, c) the value of $\lambda_1$ during optimization, d) ARE values for $\lambda_1$ during optimization.*

### 4- Discussion

The idea of using PINNs in leveraging the expressiveness of NNs in scientific machine learning has received growing attention in recent years. In this work, we assessed PINNs in predicting the out-of-training-distribution testing data. The results showed that the PINNs rarely predicted acceptably for out-of-training-distribution testing data, in line with the observations of Kim et al. (2020). The most critical question is how to select an acceptable PAF. Ideal performance was achieved with terms inspired by the analytical solution of the PDE of interest. However, in problems that we usually deal with, there are no analytical solutions, and numerical approaches are solving the problems. In general, any mathematical function involved in the workflow of the solution of the investigated PDE may be tested as a PAF.

However, there are always trade-offs between the simplicity of using general AFs and the advantages of using the PAFs. Using PAFs brings many benefits, but selecting, implementing PAFs, and training the PINNs with PAFs may be challenging, as previously discussed. In addition, using PAFs requires some scientific background and knowledge about the investigating phenomena. Since the general AFs (such as *tanh*) are somehow powerful enough to regenerate the behavior of simple functions with acceptable accuracy, the advantage of using PAFs is more impressive when the non-linearity of the PAF increases; for instance, PINNs are very weak in learning and extrapolating functions including periodic behavior. However, we realized complications when the PAFs were added to the networks. The first challenge was with the domain of the function. In general, the input to a neuron is in the domain of real numbers ($z \in \mathbb{R}$). This is while the domain of PAFs may be limited. In these cases, the training process fails. So, some actions are required to be taken to extend the domain of the function into ℝ. Also, in some cases, we observed that the chance of failure of the optimizer in the PAF-included PINNs was increased in comparison to the PINNs without PAFs. In some cases, these were due to the problems in the differentiability of the defined PAF in some parts of its domain. Also, sharp changes in the function slope curve may lead to instability problems in the model. The capability of using PAFs in reducing the number of unique solutions for a PINN is a double-edged sword. Since the number of unique solutions is lower in PAF-PINNs, finding possible solutions is more challenging for the gradient descent optimizer.

## 5- Conclusions

This work introduces a new type of activation function called Physical Activation Functions (PAFs) that improves the physics-informed neural networks (PINNs). PAF, which can be categorized as a subgroup of inductive mathematical approaches for fusing physics with NNs, is a mathematical expression that gets inspiration from the physical model of the investigated phenomena. We demonstrated the potential of PAFs for both forward and inverse problems in different cases, i.e., mechanical harmonic oscillation, Burger's equation, advection-convection phenomenon, heterogeneous diffusion process, and the Laplace equation. The below points may be concluded from the results:

- Based on examples, we showed that although the performance of PINNs was significant in the training ranges, they could not predict well on out-of-training distribution ranges.
- From a theoretical point of view, the PAFs should ideally be whole or parts of the analytical solution of the PDE being studied. In practice, analytical solutions may be unavailable, and PAFs can be obtained from analytical solutions of the PDE under more basic assumptions but same temporal-spatial behavior or be inspired by the initial or boundary conditions.
- We tried to compare the applicability of the two most prominent optimizers, i.e., Adam and LBFGS, in the training of PAF-PINN models. The results showed that both were successful in general. However, the failure cases in LBFGS were lower. In Adam optimizer, the training process was sensitive to the learning rate, while the LBFGS could work with broader learning rates. In general, we recommend the LBFGS optimizer for this purpose.
- Using PAFs may help reduce the PINN's network size by up to 75%. Furthermore, PAFs make PINNs more physically constrained. This improves prediction and reduces errors of out-of— training-distribution predictions. Additionally, using PAFs in PINNs may help reduce the training complexity of the PINNs and the final loss values. In some examples, the total loss term is reduced by 1-2 orders of magnitude.
- The PAFs are also effective in deep PINNs when PAFs are only active in a small fraction of the neurons. However, by increasing the model's width, the impact of the PAFs was reduced in the examples of this paper. Likely, the optimizer could not allocate enough importance to the neurons using PAFs.
- PAFs could improve PINN inverse calculations for the studied cases by reducing the degree of freedom of the NN model. However, over-using the PAFs did lessen the accuracy of the inverse calculations.

**Acknowledgments**

Andersen acknowledges the Research Council of Norway and the industry partners of NCS2030 – RCN project number 331644 – for their support.

**Data Availability**

After the journal version of this research is published, the codes will be accessible from the authors' GitHub repository via the link: https://github.com/jcabbasi/paf_in_pinns

**Credit author statement**

J. Abbasi: Methodology, Code Development, Derivation, Visualization, Writing; P.Ø. Andersen: Conceptualization, Methodology, Resources, Reviewing

**Declaration of competing interest**

The authors declare that they have no known competing financial interests or personal relationships that could have appeared to influence the work reported in this paper